\documentclass{ecai2010}
\usepackage{newalg}
\usepackage{graphicx}
\newcommand{\myOmit}[1]{}

\begin{document}

\title{An Empirical Study of the Manipulability of Single Transferable Voting}


\author{Toby Walsh\institute{NICTA and UNSW, Sydney, Australia,
email: toby.walsh@nicta.com.au}}

\maketitle

\begin{abstract}
Voting is a simple mechanism
to combine together the preferences of multiple
agents. Agents may try to manipulate
the result of voting by mis-reporting their preferences. 
One barrier that might exist to such manipulation is computational
complexity. In particular, it has been shown that it is NP-hard
to compute how to manipulate a number of 
different voting rules. However, NP-hardness only
bounds the worst-case complexity. Recent theoretical
results suggest that manipulation may often
be easy in practice. In this paper, we 
study empirically the manipulability of 
single transferable voting (STV) to determine
if computational complexity is really a barrier
to manipulation.
STV was one of the first voting rules
shown to be NP-hard. It also appears 
one of the harder voting rules to manipulate. 
We sample a number of 
distributions of votes including uniform and
real world elections. In almost every election in our 
experiments, it was easy to compute how a single
agent could manipulate the election or to prove
that manipulation by a single agent was impossible. 
\end{abstract}

\section{INTRODUCTION}

Agents may try to manipulate
an election by mis-reporting their preferences in
order to get a better result for themselves. 
The Gibbard Satterthwaite theorem proves that, under some
simple assumptions, there will always exist
situations where such manipulation is possible \cite{gs1,gs2}. 
In an influential paper \cite{bartholditoveytrick},
Bartholdi, Tovey and Trick 
proposed an appealing escape:
perhaps it is computationally so difficult to find a successful
manipulation that agents have little
option but to report their true preferences? 
To illustrate this idea, 
they demonstrated that the second order Copeland rule
is NP-hard to manipulate. 
Shortly after, Bartholdi and Orlin proved
that the more well known Single Transferable Voting
(STV) rule is NP-hard to manipulate \cite{stvhard}.
A whole subfield of social choice 
has since grown from this proposal, 
proving that various voting rules are NP-hard 
to manipulate under different assumptions.

Our focus here is on the manipulability of the STV rule. 
Bartholdi and Orlin argued that 
STV is one of the most promising voting
rules to consider in this respect:
\begin{quote}
``STV is apparently
unique among voting schemes in actual use today 
in that it is computationally
resistant to manipulation.'' 
(page 341 of \cite{stvhard}).
\end{quote}
Whilst there exist other voting rules which are NP-hard to manipulate,
computational complexity is either restricted to 
somewhat obscure voting rules like second order
Copeland or to more well known voting rules but with
the rather artificial restriction that there are large
weights on the votes. STV is the only commonly used voting rule
that is NP-hard to manipulate without weights. 
STV also appears more difficult to
manipulate than many other rules. For example, 
Chamberlain studied \cite{cbs85} four different measures of the manipulability
of a voting rule: the probability
that manipulation is possible,
the number of candidates who can 
be made to win, the coalition
size necessary to manipulate, and
the margin-of-error which still
results in a successful manipulation.
Compared to other commonly
used rules like plurality and Borda, 
his results showed that
STV was the most difficult to manipulate
by a substantial margin. He concluded that:
\begin{quote}
``[this] superior performance \ldots combined
with the rather complex
and implausible nature of the strategies to manipulate
it, suggest that it [the STV rule] may be quite
resitant to manipulation'' (page 203 of \cite{cbs85}).
\end{quote}

Unfortunately, the NP-hardness of manipulating 
STV is only a worst-case result and may not reflect
the difficulty of manipulation in practice. Indeed, a number of 
recent theoretical results 
suggest that manipulation can often be computationally
easy on average \cite{csaaai2006,prjair07,xcec08,fknfocs09,xcec08b}. 
Such theoretical results typically provide approximation methods so do
not say what happens with the complete methods
studied here (where worst case behaviour is exponential). 
Most recently, Walsh has suggested that 
empirical studies might provide insights
into the computational complexity of manipulation
that can complement such theoretical
results \cite{wijcai09}. 
However, Walsh's empirical study 
was 
limited to the simple veto rule, weighted votes
and elections with only three candidates. In this paper, we relax
these assumptions and 
consider the more complex multi-round STV rule, 
unweighted votes,
and large numbers of 
candidates. 



\section{MANIPULATING STV}

Single Transferable Voting
(STV) proceeds in a number of rounds.
We consider the case of electing a single winner.
Each agent totally ranks the candidates. 
Unless one candidate has a majority of
first place votes, we eliminate the candidate
with the least number of first place votes. 
Any ballots placing the eliminated candidate in first 
place are re-assigned to the second place candidate.
We then repeat until one candidate has a majority. 
STV is used in a wide variety of elections
including for the Irish presidency, the
Australian House of Representatives, the Academy awards,
and many organizations including the 
American Political Science Association,
the International Olympic Committee, 
and the British Labour Party. 

\begin{figure*}[tbh]

\begin{center}
\begin{algorithm}{Manipulate}{c,R,(s_1,\ldots,s_m),f}
\begin{IF}{|R|=1 \mbox{\it \ \ \ \ \ \ \ \ \ \ \ \ \ \ \ \ \ \ \ \ \ \ \ \ \ \ \ \ \ \ \ \ \ \ \ \ \ \ \ \ ; Is there one candidate left?}}
   \RETURN (R=\{c\}) \mbox{\it \ \  \ \ \ \ \ \ \ \ \ \ \ \ \ \ \  \ \ ; Is it the chosen candidate?}   
\end{IF} \\
\begin{IF}{f=0 \mbox{\it  \ \ \ \ \ \ \ \ \ \ \ \ \ \ \ \ \ \ \ \ \ \ \ \ \ \ \ \ \ \ \ \ \ \ \ \ \ \ \ \ \  \ ; Is the top of the manipulator's vote currently free?}} \\
d \= arg \ min_{j \in R} (s_j) 
\mbox{\it \ \ \ \ \ \ \ \ \ \ \ \ \ \ \ \ ; Who will currently be eliminated?} \\
s_d \= s_d + w 
\mbox{\it \ \ \ \ \ \ \ \ \  \ \ \ \ \ \ \ \ \ \ \ \ \ \ \  \ \ ; Suppose the manipulator votes for them} \\
e \= arg \ min_{j \in R} (s_j) \\
\begin{IF}{d = e \mbox{\it \ \ \ \ \ \ \ \ \ \ \ \ \ \ \ \ \ \ \ \ \ \ \ \ \ \ \ \ \ \ \ \ ; Does this not change the result?}} 
\RETURN \\ (c \neq d) \ and \
  \CALL{Manipulate}(c,R-\{d\},Transfer((s_1,\ldots,s_m),d,R),0)
\ELSE 
   \RETURN  \\
  ((c \neq d) \ and \ \CALL{Manipulate}(c,R-\{d\},Transfer((s_1,\ldots,s_m),d,R),0)) \ or \\
  ((c \neq e) \ and \ \CALL{Manipulate}(c,R-\{e\},Transfer((s_1,\ldots,s_m),e,R),d)) 
\end{IF}
\ELSE
\mbox{\it  \ \ \ \ \ \ \ \ \ \ \ \ \ \ \ \ \ \ \ \ \ \ \  \ \ \ \ \ \ \ \ \ \ \ \ \ \  \ \ \ \ \ ; The top of the manipulator's vote is fixed} \\
d \= arg \ min_{j \in R} (s_j) 
\mbox{\it \ \ \ \ \ \ \ \ \ \ \ \ \ \ \ \ \ ; Who will be eliminated?} \\
\begin{IF}{c = d \mbox{\it \ \ \ \ \ \ \ \ \ \ \ \ \ \ \ \ \ \ \ \ \ \ \ \ \ \ \ \ \ \ \ \ \ ; Is this the chosen candidate?}} {\RETURN false}
\end{IF} \\
\begin{IF}{d = f \mbox{\it \ \ \ \  \ \ \ \ \ \ \ \ \ \ \ \ \ \ \ \ \ \ \ \ \ \ \ \ \ \ \ \ ; Is the manipulator free again to change the result?}} 
   \RETURN  
  \CALL{Manipulate}(c,R-\{d\},Transfer((s_1,\ldots,s_m),d,R),0) 
\ELSE
   \RETURN  
  \CALL{Manipulate}(c,R-\{d\},Transfer((s_1,\ldots,s_m),d,R),f) 
\end{IF}
\end{IF}
\end{algorithm}
\end{center}

\caption{Our improved algorithm to compute
if an agent can manipulate a STV election. }
\vspace{-2em}
{
\scriptsize We 
use integers from $1$ to $m$ for
the candidates, 
integers from $1$ to $n$ for
the agents (with $n$ being the manipulator),
$c$ for the candidate who the manipulator wants to win,
$R$ for the set of un-eliminated
candidates, $s_j$ for the weight
of agents ranking candidate $j$ first
amongst $R$, $w$ for the weight of 
the manipulator, and $f$ for the candidate most highly
ranked by the manipulator amongst $R$ 
(or $0$ if there is currently no constraint
on who is most highly ranked). 
The function $Transfer$
computes the a vector of the new weights
of agents ranking candidate $j$ first
amongst $R$ after a given candidate
is eliminated.
The algorithm is initially called with
$R$ set to every candidate, and
$f$ to $0$.  }
\label{fig-alg}
\end{figure*}

STV is NP-hard to manipulate
by a single agent if the number of candidates is
unbounded and votes are unweighted
\cite{stvhard}, or by a coalition of agents
if there are 3 or more candidates and
votes are weighted 
\cite{csljacm07}.
Coleman and Teague
give an enumerative method
for a coalition of $k$ unweighted
agents to compute a
manipulation of the
STV rule which runs
in $O(m!(n+mk))$ time 
where $n$ is the number of agents voting and $m$ is
the number of candidates \cite{ctcats2007}.
For a single manipulator, 
Conitzer, Sandholm and Lang give an 
$O(n 1.62^m)$ time algorithm
(called CSL from now on)
to compute the set of candidates that 
can win a STV election
\cite{csljacm07}. 

In Figure \ref{fig-alg}, we give a new
algorithm for computing a manipulation 
of the STV rule which 
improves upon CSL in several directions
First, our algorithm ignores
elections in which the chosen candidate is eliminated.
Second, our algorithm terminates search as soon as 
a manipulation is found in which the chosen
candidate wins. Third, our algorithm does not explore the
left branch of the search tree when the
right branch gives a successful manipulation.

\begin{table}
\begin{center}
  \begin{tabular}{| r || r | r | r | r | } 
  \hline
    & \multicolumn{2}{|c|}{CSL algorithm} & 
\multicolumn{2}{|c|}{Improved algorithm} \\ 
$n$ & nodes & time/s & nodes & time/s  \\ \hline
2 & 1.46 & {\bf 0.00} & {\bf 1.24} & {\bf 0.00} \\
4 & 3.28 & {\bf 0.00} & {\bf 1.59} & {\bf 0.00} \\
8 & 11.80 & {\bf 0.00} & {\bf 3.70} & {\bf 0.00} \\
16 & 59.05 & 0.03 & {\bf 12.62} & {\bf 0.01} \\
32 & 570.11 & 0.63 & {\bf 55.20} & {\bf 0.09} \\
64 & 14,676.17 & 33.22 & {\bf 963.39} & {\bf 3.00} \\
128 & 8,429,800.00 & 6,538.13 & {\bf 159,221.10} & {\bf 176.68} \\ \hline
  \end{tabular}
\end{center}
\caption{Comparison between the CSL algorithm
and our improved algorithm to compute
a manipulation of a STV election. 
}
\label{table-alg}
\end{table}

To show the benefits of these improvements,
we ran an experiment in which $n$ agents
vote uniformly at random over $n$ possible candidates.
The experiment was run 
in CLISP 2.42 on a 3.2 GHz Pentium 4 with 3GB of memory
running Ubuntu 8.04.3.
Table \ref{table-alg}. 
gives
the mean nodes explored and runtime needed to compute a 
manipulation or prove none exists. Median and other
percentiles display similar behaviour. 
We see that our new method can be more than an order
of magnitude faster than CSL. 
In addition, as problems
get larger, the improvement increases.
At $n=32$, our 
method is nearly 10 times faster than CSL. This increases to 
roughly 40 times
faster at $n=128$. These improvements reduce
the time to find a manipulation on the largest
problems from several hours to a couple of minutes. 

\section{UNIFORM VOTES}

We start with one of the simplest possible
scenarios: elections in which each vote is equally
likely. We have one agent trying to manipulate an
election of $m$ candidates where $n$ other
agents vote. Votes are drawn uniformly
at random from all $m!$ possible votes. 
This is the Impartial Culture (IC) model.

\subsection{VARYING THE AGENTS}

\begin{figure}[htb]
\vspace{-2.5in}
\begin{center}
\includegraphics[scale=0.4]{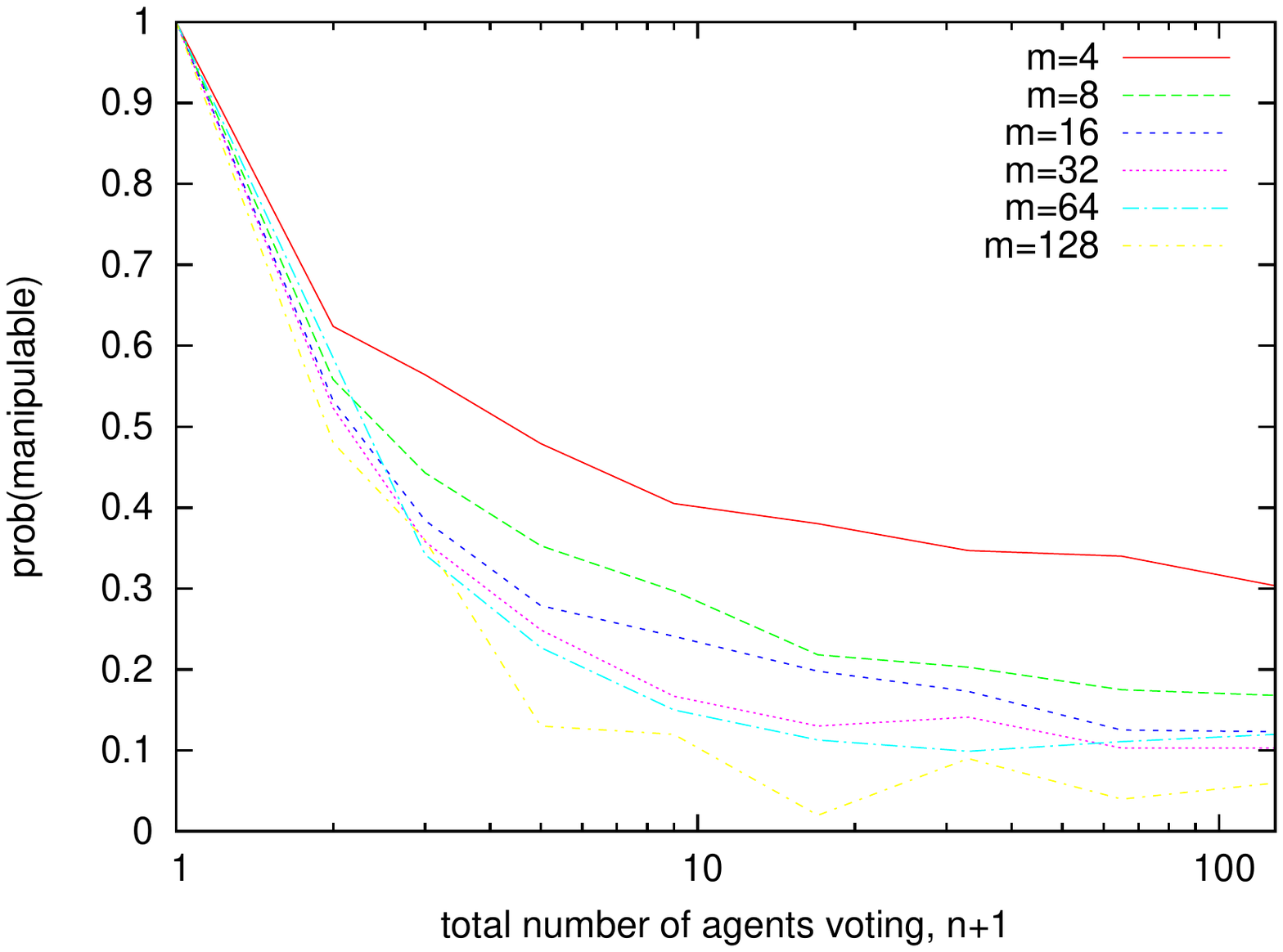}
\end{center}
\vspace{-0.5in}
\caption{Manipulability of random uniform votes.
The number of candidates is fixed and we vary the number of
agents. 
}
\label{fig-prob-varn}
\end{figure}

\begin{figure}[htb]
\vspace{-2.5in}
\begin{center}
\includegraphics[scale=0.4]{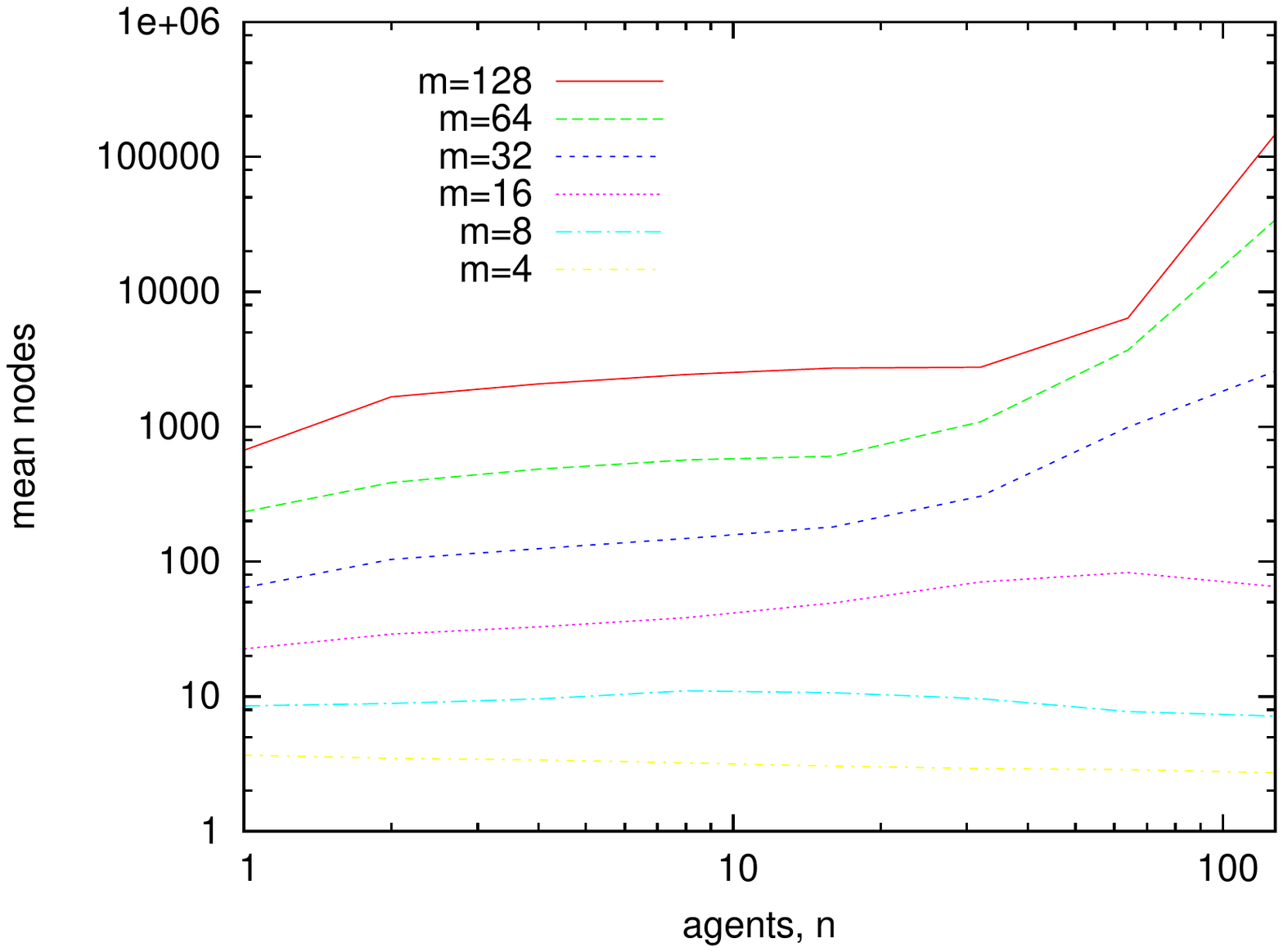}
\end{center}
\vspace{-0.5in}
\caption{Search to compute if an agent can manipulate an election
with random uniform votes.
The number of candidates is fixed and we vary the number of
agents. 
}
\label{fig-nodes-varn}
\end{figure}

In Figures \ref{fig-prob-varn} and \ref{fig-nodes-varn},
we plot the probability that a manipulator can 
make a random agent win, and the cost to compute
if this is possible when we fix the number of
candidates but vary the number of agents
in the election. 
In this and subsequent experiments,
we tested 1000 problems at each point.
Unless otherwise indicated, 
the number of candidates and 
of agents are varied in powers of 2 from
1 to 128.

The ability of an agent
to manipulate the election decreases as the
number of agents increases. 
Only if there are few votes and 
few candidates is there a significant chance
that the manipulator will be able to change
the result. 
Unlike 
domains
like satisfiability \cite{mitchell-hard-easy,SAT-phase},
constraint satisfaction \cite{gmpwcp95,random},
number partitioning \cite{rnp,gw-ci98} and
the traveling salesperson problem \cite{GentIP:tsppt}, 
the probability curve does not appear to sharpen to 
a step function around a fixed point. The probability
curve resembles the smooth phase transitions
seen in polynomial problems
like 2-coloring \cite{achlioptasphd} and 1-in-2 satisfiability
\cite{waaai2002}. 
Note that as elsewhere, we assume that ties
are broken in favour of the manipulator.
For this reason, the probability that an election
is manipulable is greater than $\frac{1}{m}$. 

Finding a manipulation or proving none
is possible is easy unless we have both a 
a large number of agents and a large number
of candidates. However, in this situation, 
the chance that the manipulator can
change the result is very small.

\subsection{VARYING THE CANDIDATES}

\begin{figure}[htb]
\vspace{-2.5in}
\begin{center}
\includegraphics[scale=0.4]{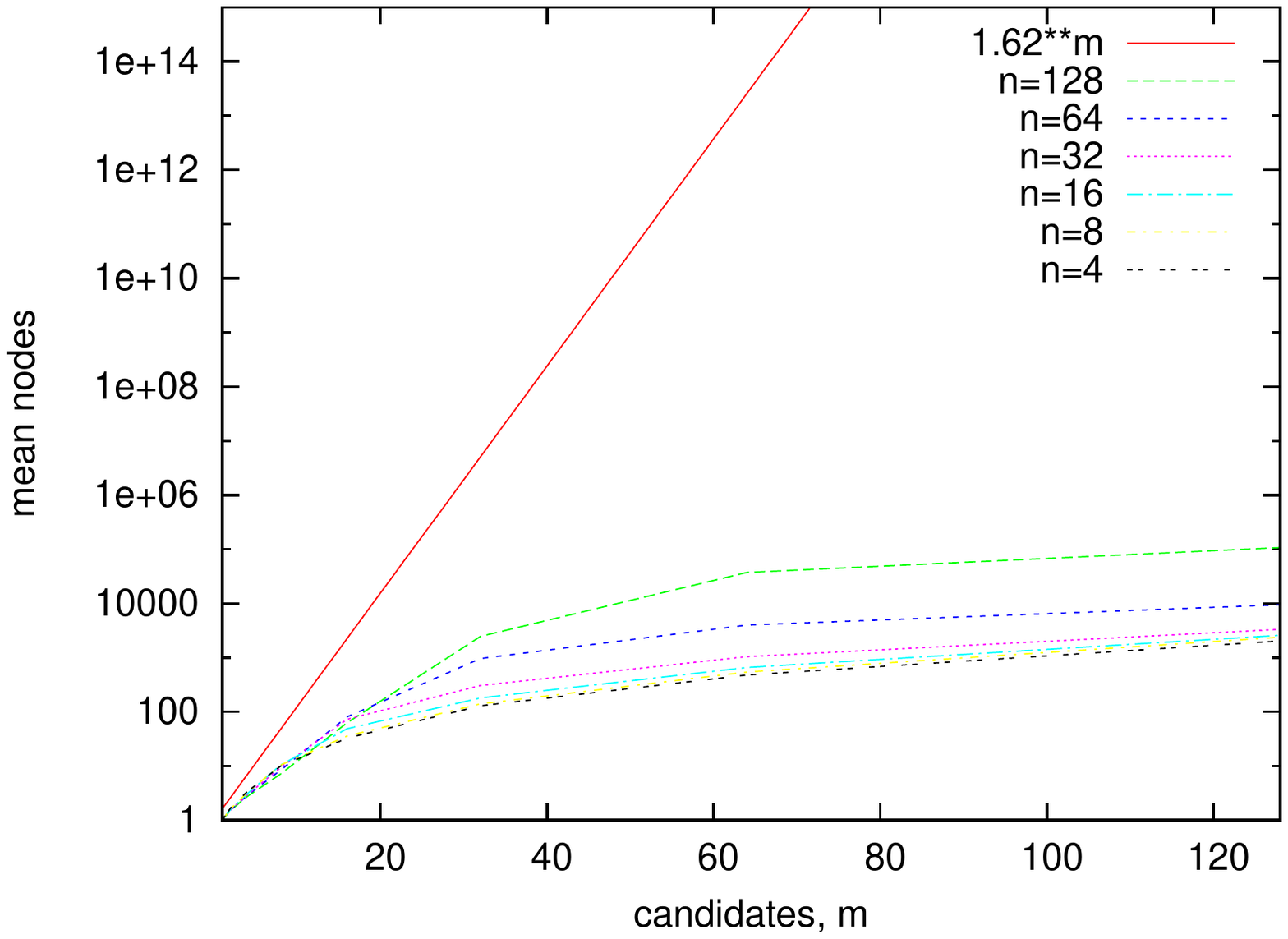}
\end{center}
\vspace{-0.5in}
\caption{Search to compute if an agent can manipulate an election
with random uniform voting.
The number of agents 
is fixed and we vary the number of
candidates. 
}
\label{fig-nodes-varm}
\end{figure}

In Figures 
\ref{fig-nodes-varm},
we plot the search to compute if the manipulator can 
make a random agent win when we 
fix the number of agents but vary the number of candidates. 
The probability curve that the manipulator can make
a random agent win resembles Figure \ref{fig-prob-varn}. 

\myOmit{
\begin{figure}[htb]
\vspace{-2.5in}
\begin{center}
\includegraphics[scale=0.4]{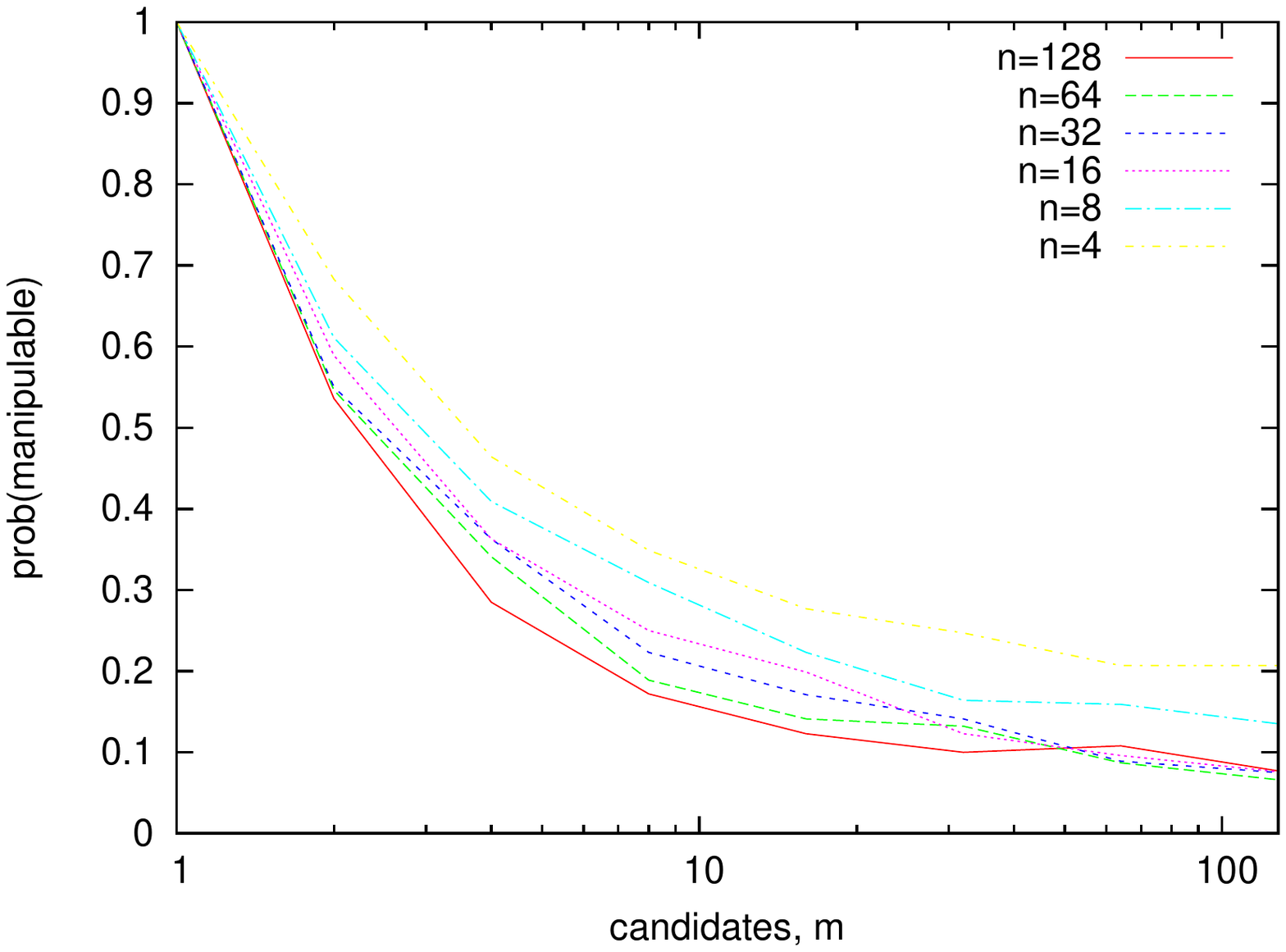}
\end{center}
\vspace{-0.5in}
\caption{Manipulability of 
random uniform voting. 
The number of agents 
is fixed and we vary the number of
candidates.
}
\label{fig-prob-varm}
\end{figure}
}

Whilst the cost of computing a manipulation
increases exponential with the number
of candidates $m$, the observed scaling is much better than
the $1.62^m$. 
We can easily compute manipulations
for up to 128 candidates. 
Note that $1.62^m$ is over $10^{26}$ for $m=128$. Thus, we
appear to be far from the worst case. 
We fitted the observed data to the model $ab^m$ and found
a good fit with $b=1.008$ and a coefficient of determination, $R^2=0.95$.

\section{URN MODEL}

In many real life situations, votes are not completely uniform
and uncorrelated with each other. What happens if 
we introduce correlation between votes? Here we consider
random votes
drawn from the Polya Eggenberger urn model \cite{polya-urn}.
We also observed very similar results when votes 
are drawn at random which are single peaked or single troughed.
In the urn model, 
we have an urn containing all $m!$ possible votes. 
We draw votes out of the urn at random, and
put them back into the urn with $a$ 
additional votes of the same type (where
$a$ is a parameter).
As $a$ increases, there is 
increasing correlation between the votes. 
This generalizes both the Impartial Culture
model ($a=0$) and the Impartial Anonymous Culture ($a=1$) model. 
To give a parameter independent of problem 
size, we consider $b=\frac{a}{m!}$. 
For instance, with $b=1$, there is a 50\% chance
that the second vote is the same as the first. 

\begin{figure}[htb]
\vspace{-2.5in}
\begin{center}
\includegraphics[scale=0.4]{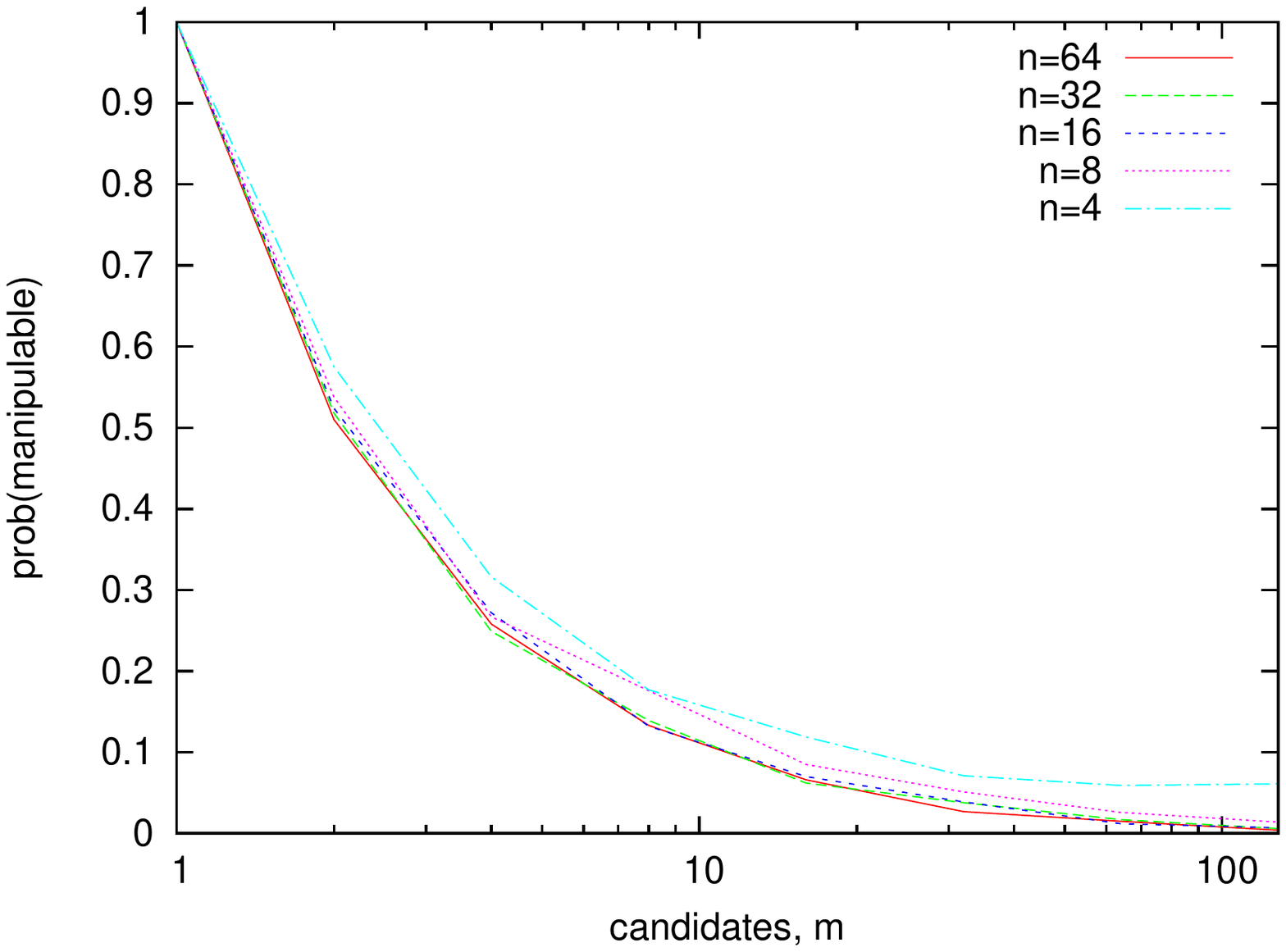}
\end{center}
\vspace{-0.5in}
\caption{Manipulability of 
correlated votes. 
The number of agents 
is fixed and we vary the number of
candidates. 
The $n$ fixed votes are drawn from the Polya Eggenberger urn  model
with $b=1$. 
}

\label{fig-urn-prob-varm}
\end{figure}
\begin{figure}[htb]
\vspace{-2.5in}
\begin{center}
\includegraphics[scale=0.4]{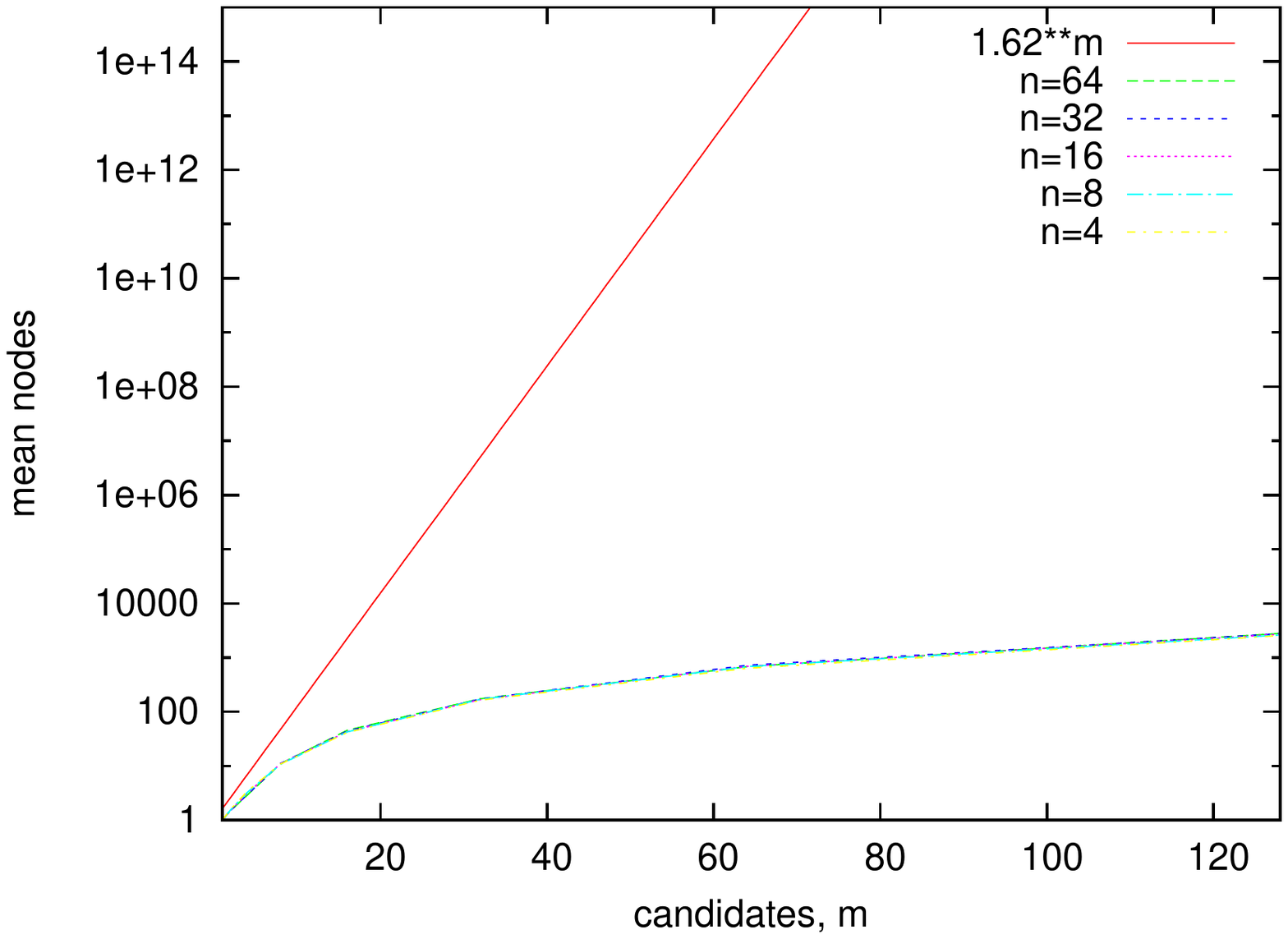}
\end{center}
\vspace{-0.5in}
\caption{Search to compute if an agent can manipulate an election
with correlated votes. 
The number of agents 
is fixed and we vary the number of
candidates. 
The $n$ fixed votes are drawn using the Polya Eggenberger urn  model
with $b=1$. 
The curves for different $n$ fit closely on top of each other.
}
\label{fig-urn-nodes-varm}
\end{figure}

\begin{figure}[htb]
\vspace{-2.5in}
\begin{center}
\includegraphics[scale=0.4]{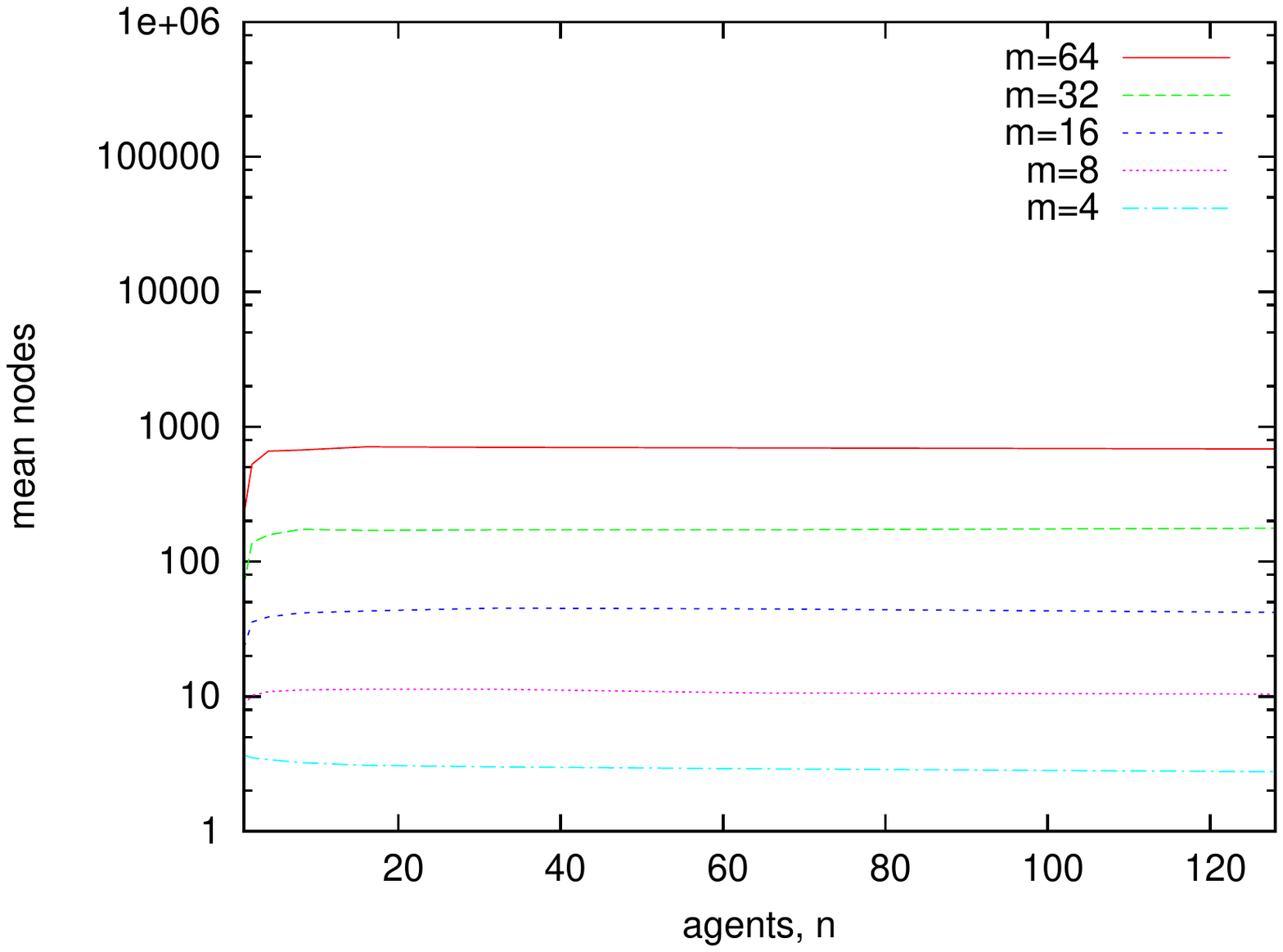}
\end{center}
\vspace{-0.5in}
\caption{Search to compute if an agent can manipulate an election
with correlated votes. 
The number of candidates is fixed and we vary the number of
agents. 
The $n$ fixed votes are drawn using the Polya Eggenberger urn  model
with $b=1$. 
}
\label{fig-urn-nodes-varn}
\end{figure}

In Figures \ref{fig-urn-prob-varm} and \ref{fig-urn-nodes-varm},
we plot the probability that a manipulator can 
make a random agent win, and the cost to compute
if this is possible as we vary the number of candidates
in an election where votes are 
drawn from the Polya Eggenberger urn model.
The search cost to compute a manipulation
increases exponential with the number
of candidates $m$. However, we can
easily compute manipulations
for up to 128 candidates and agents. 
We fitted the observed data to the model $ab^m$ and found
a good fit with $b=1.001$ and 
a coefficient of determination, $R^2=0.99$.

In Figure \ref{fig-urn-nodes-varn},
we plot the cost to compute a manipulation
when we fix the number of candidates
but vary the number of agents. 
As in previous experiments, finding a manipulation or proving none 
exists is easy even if we have many 
agents and candidates.
We also saw very similar results
when we generated single peaked votes
using an urn model.

\section{COALITION MANIPULATION}

\begin{figure}[htb]
\vspace{-2.5in}
\begin{center}
\includegraphics[scale=0.4]{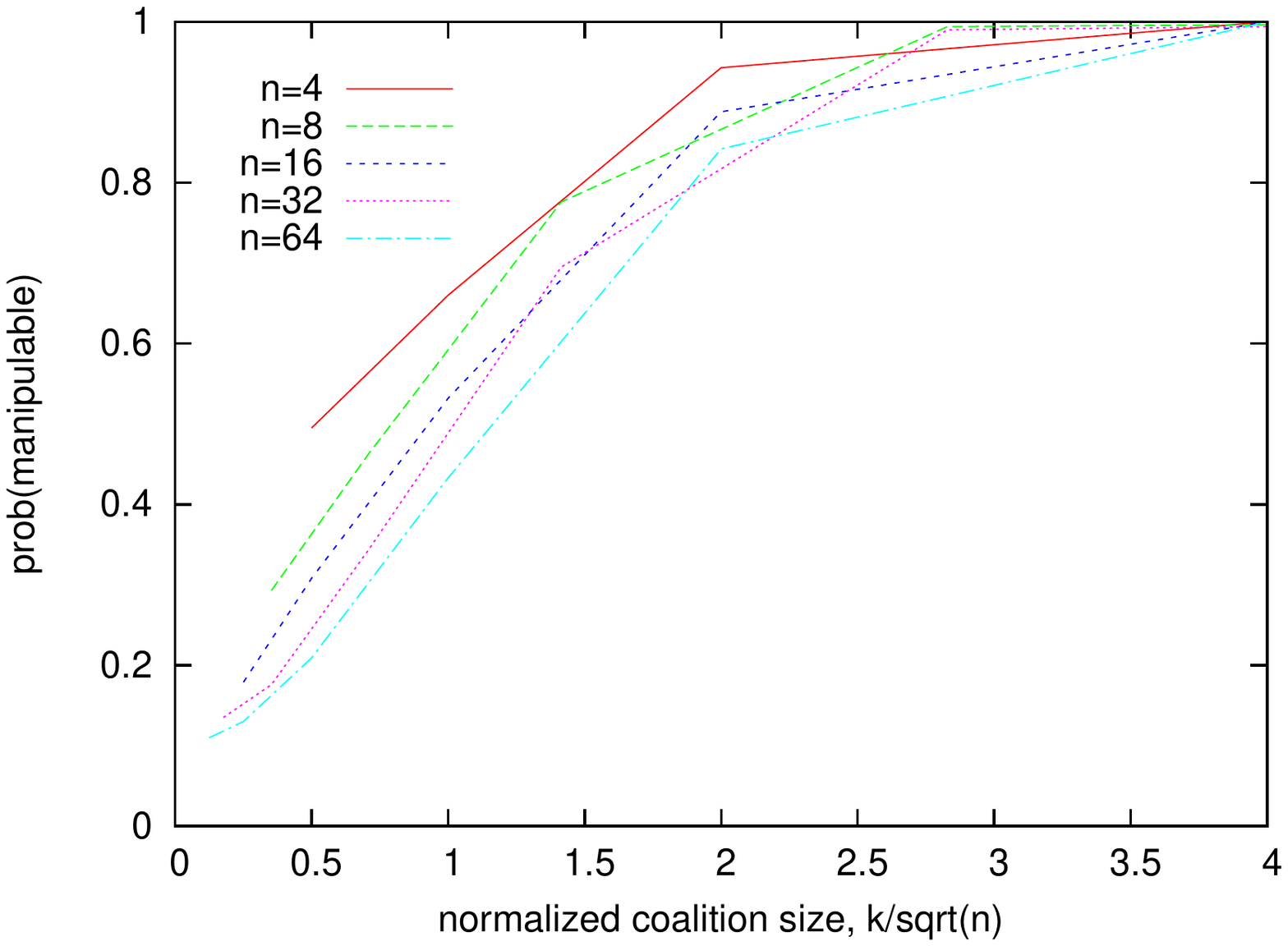}
\end{center}
\vspace{-0.5in}
\caption{Manipulability of an election
as we vary the size of the manipulating coalition. 
The number of candidates is the same
as the number of non-manipulating agents. 
}
\label{fig-prob-varw}
\end{figure}

\begin{figure}[htb]
\vspace{-2.5in}
\begin{center}
\includegraphics[scale=0.4]{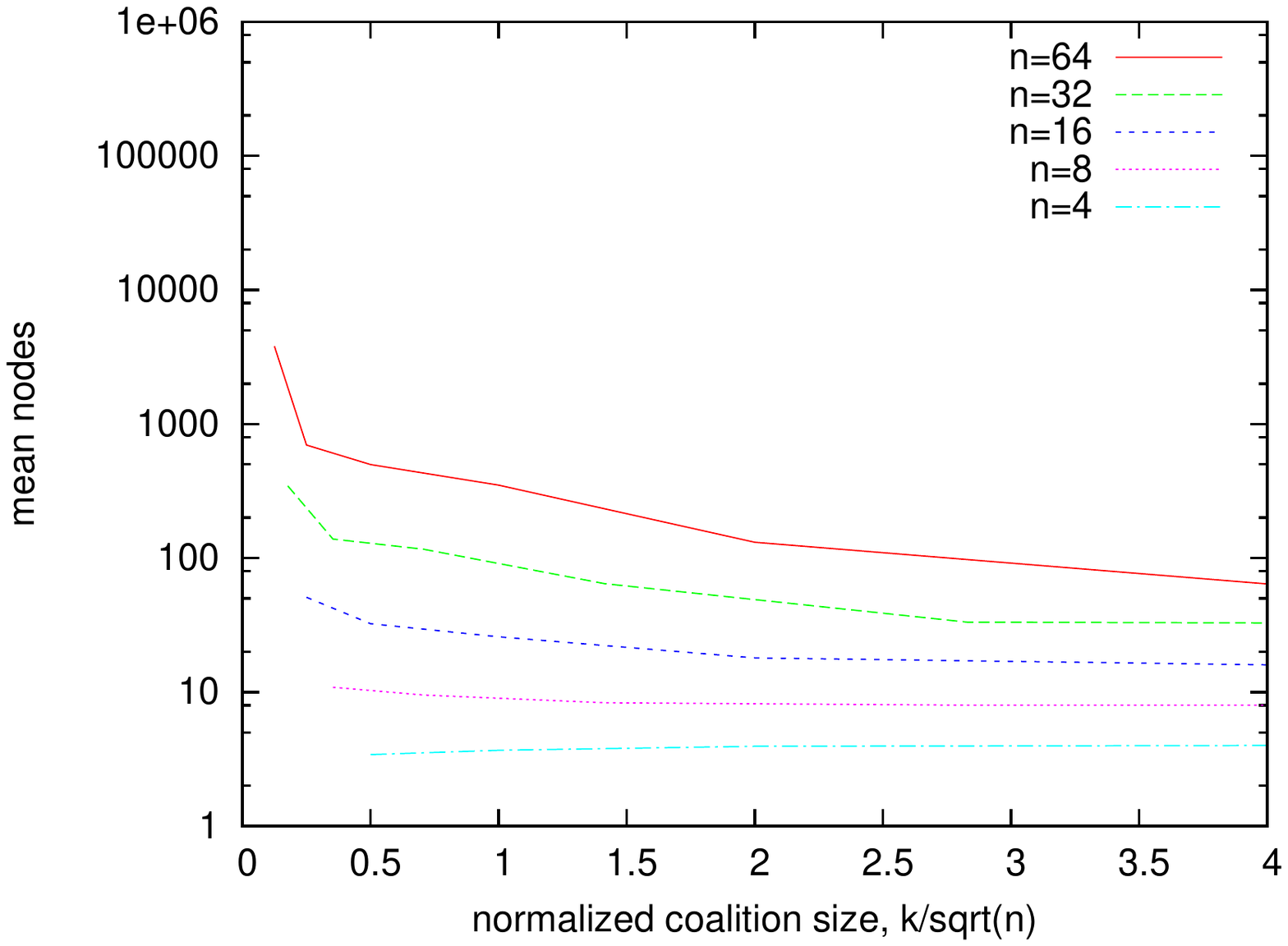}
\end{center}
\vspace{-0.5in}
\caption{Search to compute if a coalition can manipulate an election
as we vary coalition size. 
}
\label{fig-nodes-varw}
\end{figure}

Our algorithm for computing manipulation by a single agent can also
be used to compute if a coalition can manipulate 
an election when the members of coalition vote
in unison. This ignores more
complex manipulations where the members of the coalition
need to vote in different
ways. Insisting that the members of the coalition vote in unison
might be reasonable
if we wish manipulation to have both a low computational
and communication cost. 
In Figures \ref{fig-prob-varw} and \ref{fig-nodes-varw},
we plot the probability that a coalition voting in unison can 
make a random agent win, and the cost to compute
if this is possible as we vary the size of the coalition. 
Theoretical results in \cite{xcec08} and elsewhere suggest that
the critical size of a coalition that can just manipulate
an election grows as $\sqrt{n}$. We therefore normalize
the coalition size by $\sqrt{n}$. 

The ability of the coalition
to manipulate the election increases as the
size of the coalition increases. 
When the coalition is about $\sqrt{n}$ in size,
the probability that the coalition can manipulate
the election so that a candidate chosen at random wins 
is around $\frac{1}{2}$.  
The cost to compute a manipulation (or prove
that none exists) decreases as we increase the size of
the coalition. It is easier for a larger coalition 
to manipulate an election than
a smaller one. 

These experiments again suggest
different behaviour occurs here
than in other combinatorial problems like
propositional satisfiability and graph colouring 
\cite{cheeseman-hard,waaai98,wijcai99,wijcai2001}. 
For instance, we do not see a rapid transition
that sharpens around a fixed point
as in 3-satisfiability \cite{mitchell-hard-easy}. 
When we vary the
coalition size, we see a transition in the probability
of being able to manipulate the result around
a coalition size $k=\sqrt{n}$. However, this transition
appears smooth and does not seem to sharpen 
towards a step function as $n$ increases. 
In addition, hard instances do not
occur around $k=\sqrt{n}$. Indeed, the hardest
instances are when the coalition is smaller
than this and has only a small chance of being able to manipulate
the result. 

\section{SAMPLING REAL ELECTIONS}

Elections met in practice may differ
from those sampled so far. There might, for instance, 
be some votes which are never cast. 
On the other hand, with the models studied so far every possible
random/single peaked
vote has a non-zero probability of being seen. 
We therefore sampled some real voting
records \cite{isai95,ghpwaaai99}. 

\myOmit{
\begin{figure}[htb]
\vspace{-2.5in}
\begin{center}
\includegraphics[scale=0.4]{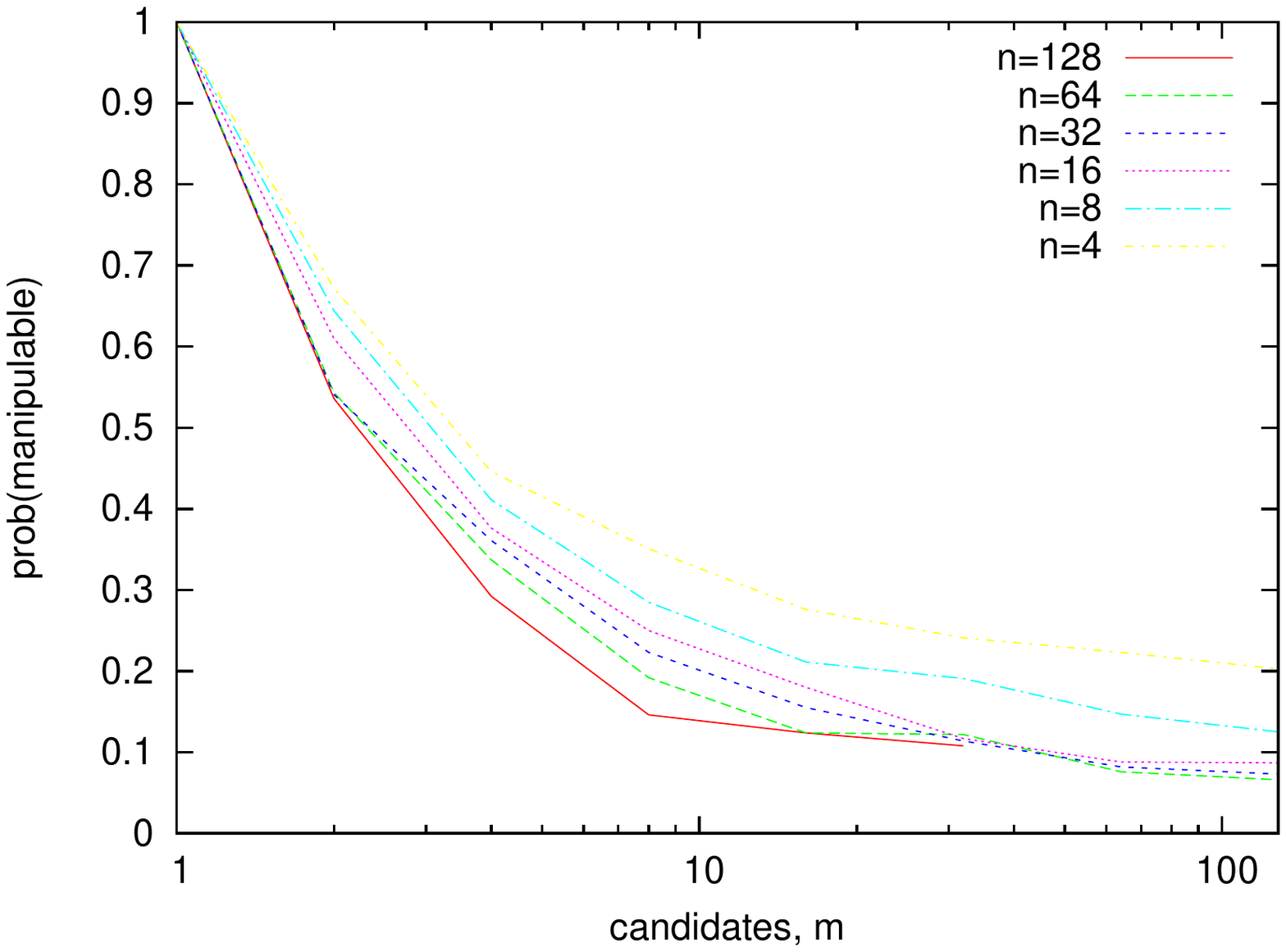}
\end{center}
\vspace{-0.5in}
\caption{Manipulability of 
votes sampled from the NASA experiment. 
The number of agents 
is fixed and we vary the number of
candidates. 
}
\label{fig-nasa-prob-varm}
\end{figure}
}

\begin{figure}[htb]
\vspace{-2.5in}
\begin{center}
\includegraphics[scale=0.4]{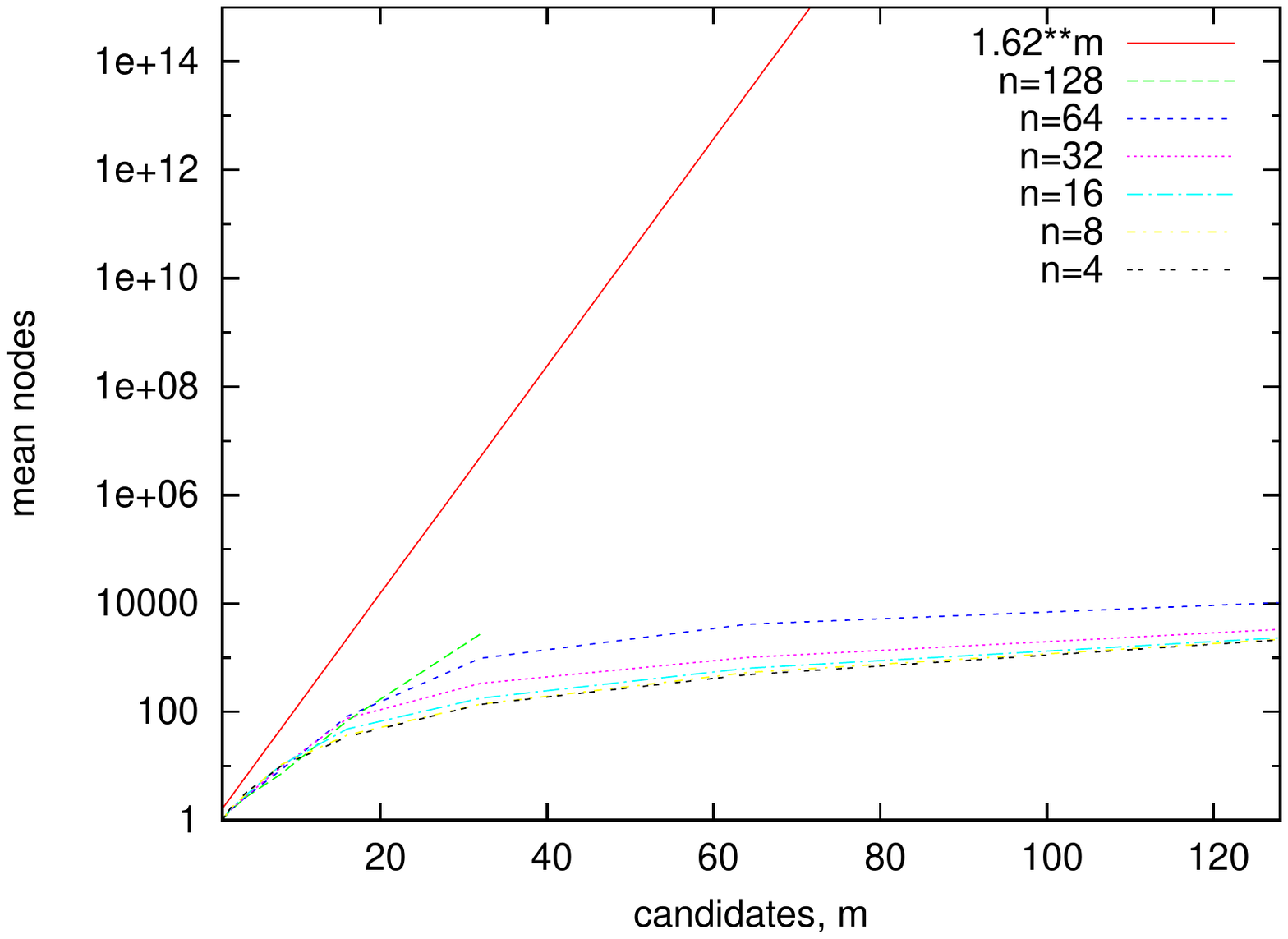}
\end{center}
\vspace{-0.5in}
\caption{Search to compute if an agent can manipulate an election
with votes sampled from the NASA experiment. 
The number of agents 
is fixed and we vary the number of
candidates. 
}
\label{fig-nasa-nodes-varm}
\end{figure}

\begin{figure}[htb]
\vspace{-2.5in}
\begin{center}
\includegraphics[scale=0.4]{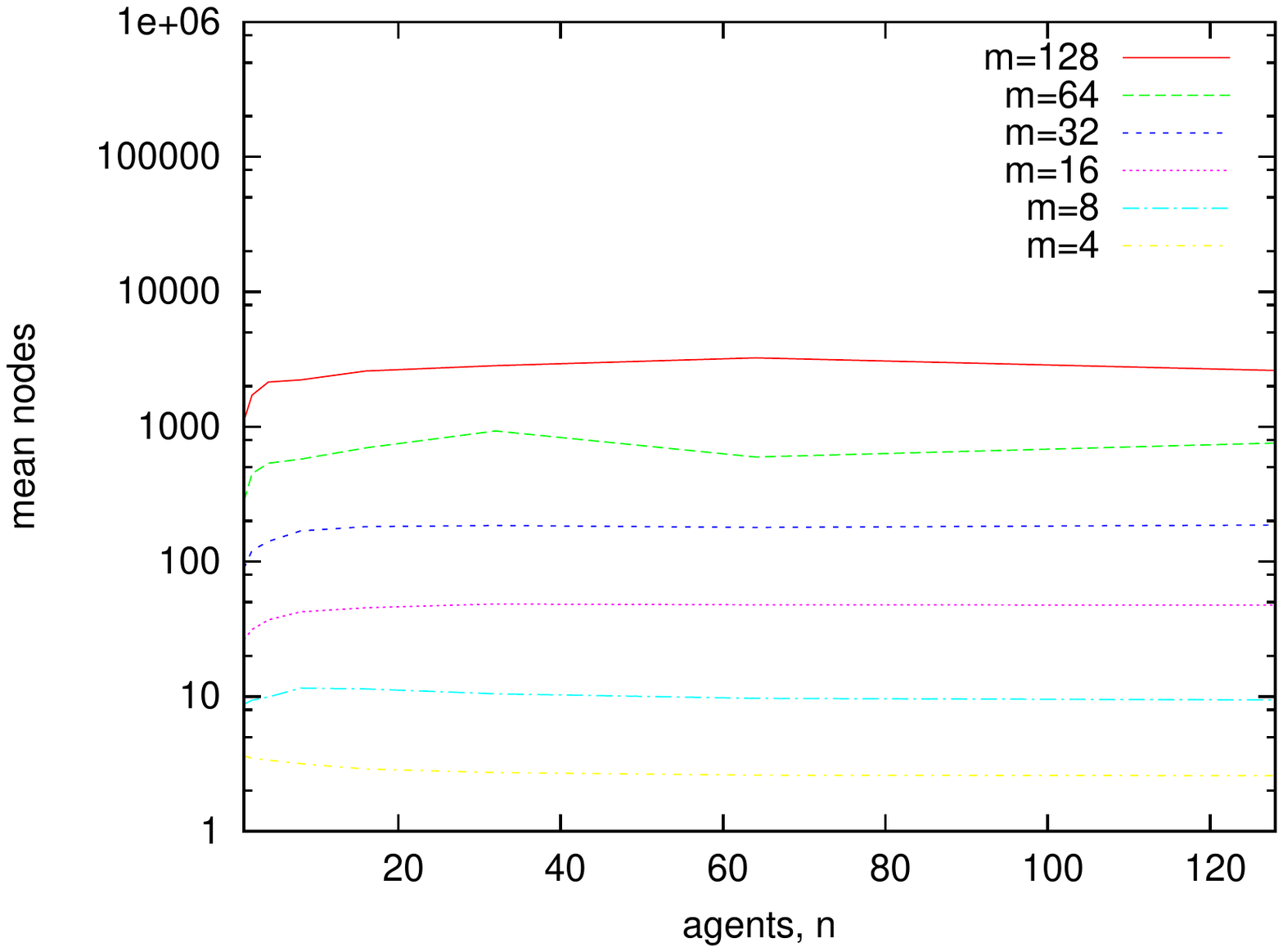}
\end{center}
\vspace{-0.5in}
\caption{Search to compute if an agent can manipulate an election
with votes sampled from the NASA experiment. 
The number of candidates is fixed and we vary the number of
agents. 
}
\label{fig-nasa-nodes-varn}
\end{figure}

Our first experiment uses the votes cast by 10 teams 
of scientists to select one of 32 different
trajectories for NASA's Mariner
spacecraft \cite{mariner}. Each team ranked the different
trajectories based on their scientific value. 
We sampled these votes. 
For elections with 10 or fewer agents voting,
we simply took a random subset of the 10 votes. 
For elections with more than 10 agents voting,
we simply sampled from the 10 votes with uniform
frequency. For elections with 32 or
fewer candidates, we simply took a random subset of
the 32 candidates. Finally for elections with more than 32
candidates, we duplicated each candidate and assigned
them the same ranking. Since STV works on total
orders, we then forced each agent to break any ties
randomly.

In Figures 
\ref{fig-nasa-nodes-varm} to \ref{fig-nasa-nodes-varn},
we plot 
the cost to compute
if a manipulator can make a random agent win
as we vary the number of candidates
and agents.
Votes are sampled from the NASA experiment as explained
earlier. 
The probability that the manipulator can manipulate
the election resembles the probability curve
for uniform random votes. 
The search needed to compute a manipulation
again increases exponential with the number
of candidates $m$. However, 
the observed scaling is much better than
$1.62^m$. We can
easily compute manipulations
for up to 128 candidates and agents. 

\myOmit{
\begin{figure}[htb]
\vspace{-2.5in}
\begin{center}
\includegraphics[scale=0.4]{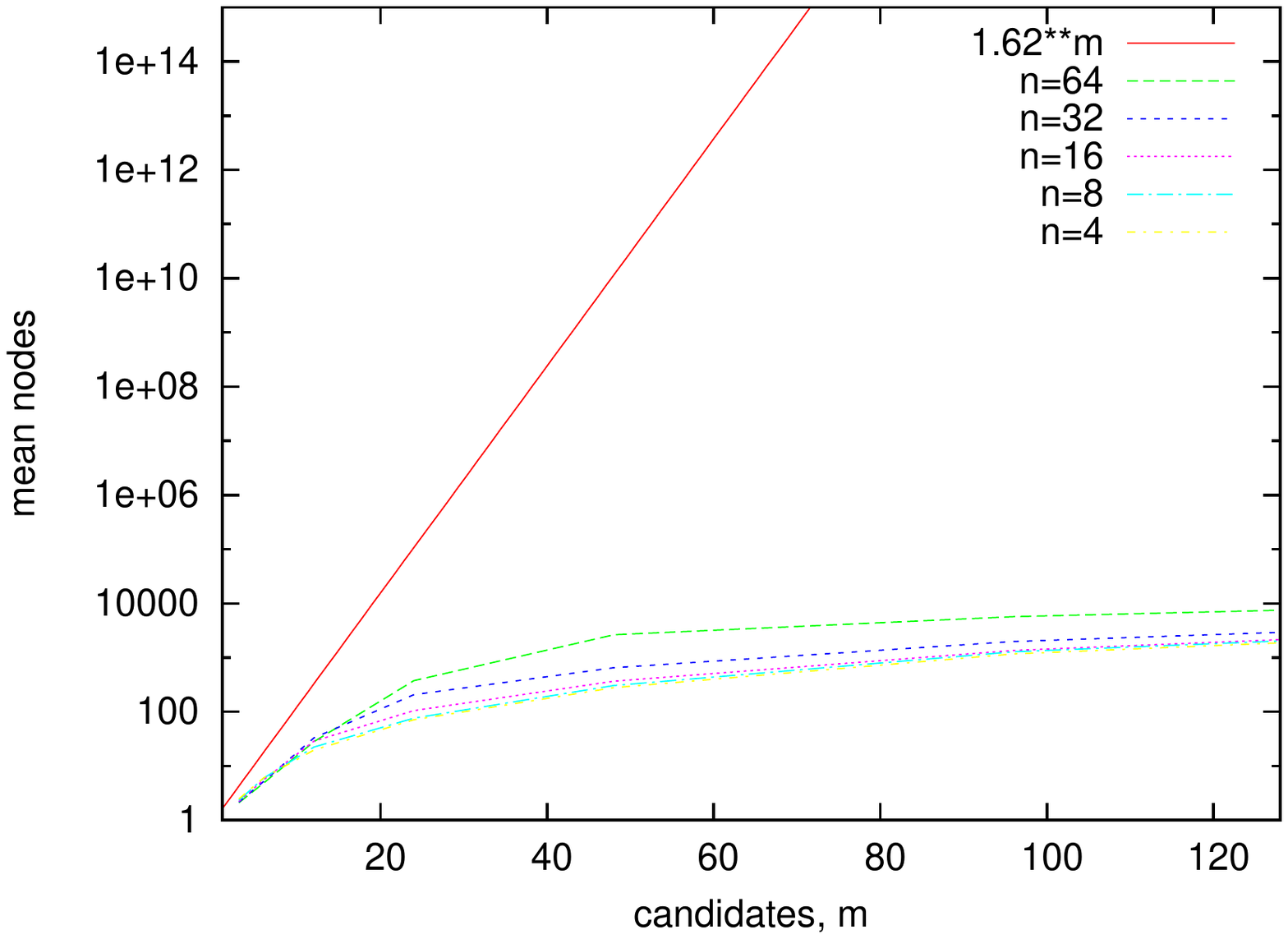}
\end{center}
\vspace{-0.5in}
\caption{Search to compute if an agent can manipulate an election
with votes sampled from a faculty hiring committee.}
{The number of agents voting is fixed and we vary the number of
candidates. 
The y-axis measures the mean 
number of search nodes explored to compute
a manipulation or prove that none exists. Median
and other percentiles are similar.}
\label{fig-dept-nodes-varm}
\end{figure}

\begin{figure}[htb]
\vspace{-2.5in}
\begin{center}
\includegraphics[scale=0.4]{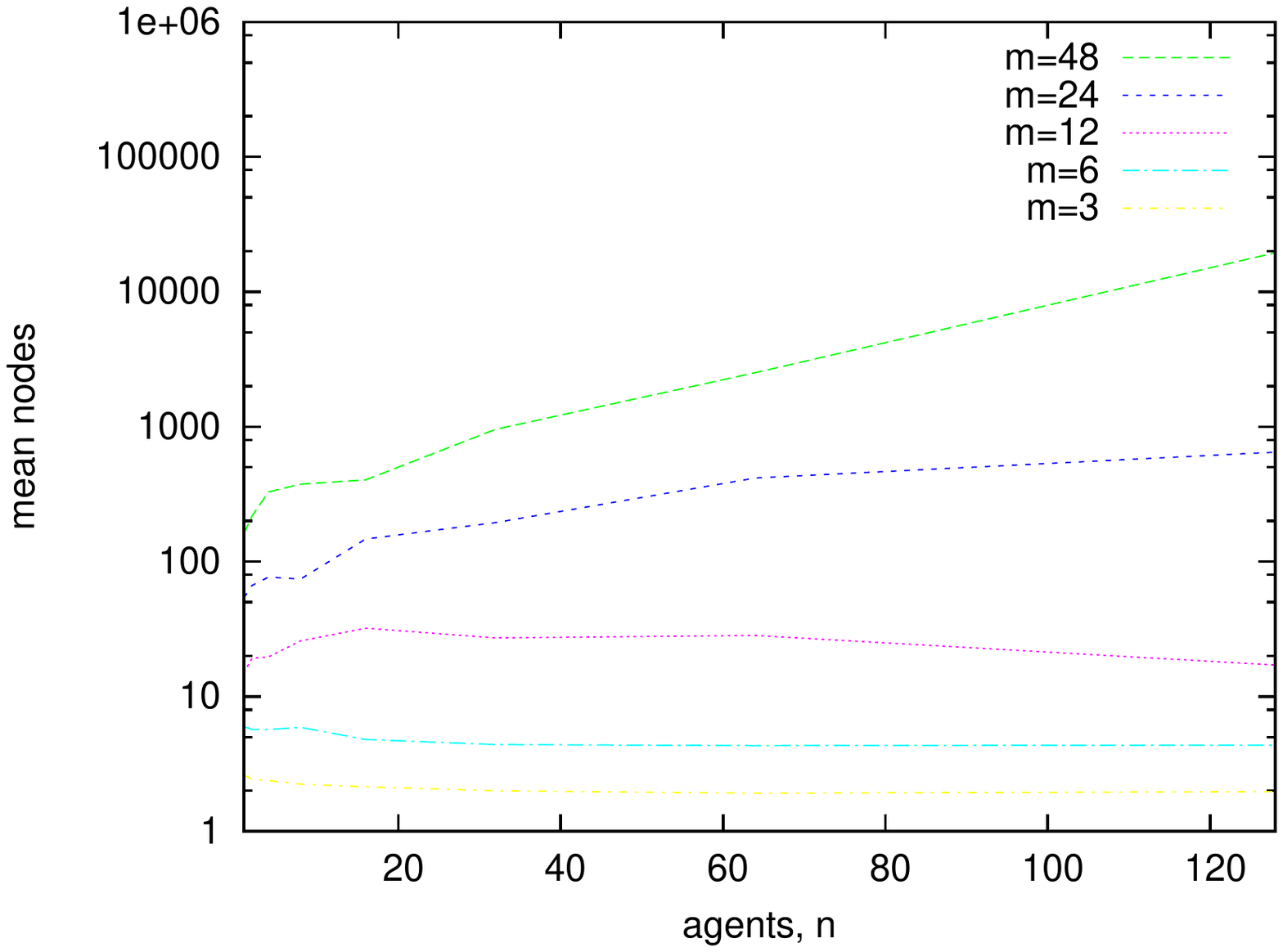}
\end{center}
\vspace{-0.5in}
\caption{Search to compute if an agent can manipulate an election
with votes sampled from a faculty hiring committee.}
{The number of candidates is fixed and we vary the number of
agents voting. 
The y-axis measures the mean 
number of search nodes explored to compute 
a manipulation or prove that none exists. Median
and other percentiles are similar.}
\label{fig-dept-nodes-varn}
\end{figure} }

In our second experiment, we used
votes from a faculty hiring committee at the University
of California at Irvine \cite{dpc83}. 
We sampled from this data set
in the same ways as from the NASA dataset
and observed very similar results. 
It was easy to find a manipulation 
or prove that none exists. 
The observed scaling was again much better than
$1.62^m$.


\section{RELATED WORK}

As indicated, there have been several theoretical
results recently that suggest elections are 
easy to manipulate in practice despite
NP-hardness results. 
For example, Procaccia and Rosenschein proved that for most
scoring rules and a wide variety of
distributions over votes, 
when the size of the coalition is $o(\sqrt{n})$, the probability
that they can change the result 
tends to 0, and when it is $\omega(\sqrt{n})$, the probability
that they can manipulate the result
tends to 1 \cite{praamas2007}.
They also gave a simple
greedy procedure that will find a manipulation 
of a scoring rule in polynomial
time with a probability of failure that is 
an inverse polynomial in $n$
\cite{prjair07}.

As a second example, Xia and Conitzer have shown that for a large class
of voting rules including STV, as the number of agents grows, either 
the probability that a coalition
can manipulate the result is very small (as the
coalition is too small), or the
probability that they can easily manipulate the result
to make any alternative
win is very large \cite{xcec08}. 
They left open only a small interval in the size
for the coalition for which the coalition is 
large enough to manipulate but
not obviously large enough to manipulate the
result easily. 

Friedgut, Kalai and Nisan proved that
if the voting rule is neutral and
far from dictatorial and there
are 3 candidates then there exists
an agent for whom a random manipulation 
succeeds with probability $\Omega(\frac{1}{n})$
\cite{fknfocs09}. 
Starting from different assumptions, Xia and Conitzer showed that
a random
manipulation would succeed with probability
$\Omega(\frac{1}{n})$ for 3 or more 
candidates for STV, for 4 or more candidates for any scoring
rule and for 5 or more candidates for Copeland \cite{xcec08b}. 

Walsh empirically studied 
manipulation of the veto rule by
a coalition of agents whose votes 
were weighted \cite{wijcai09}. 
He showed that there was a smooth transition in the probability
that a coalition can 
elect a desired candidate as the size of 
the manipulating coalition increases.
He also showed that it was easy 
to find manipulations of the veto rule 
or prove that none exist 
for many independent and identically distributed votes
even when the
coalition was critical in size. 
He was able to identify a situation
in which manipulation was computationally hard. 
This is when votes are highly correlated
and the election is ``hung''. 
However, even a single uncorrelated agent was enough to 
make manipulation easy again. 

Coleman and Teague proposed 
algorithms to compute a manipulation for
the STV rule \cite{ctcats2007}.
They also conducted an empirical study
which demonstrates that only relatively small
coalitions are needed to change the elimination
order of the STV rule. They observed that most uniform
and random elections
are not trivially manipulable using a simple greedy
heuristic. On the other hand, our results suggest 
that, for manipulation by a single agent, a limited amount of 
backtracking is needed to find
a manipulation or prove that none exists.

\section{CONCLUSIONS}

We have studied empirically
whether computational
complexity is a barrier to the manipulation
for the STV rule.
We have looked at a number of different
distributions of votes including
uniform random votes,
correlated votes drawn from an
urn model, and votes sampled from some
real world elections.
We have looked at manipulation 
by both a single agent, and a coalition
of agents who vote in unison.
Almost every one of the millions of
elections in our 
experiments was easy to manipulate or to prove
could not be manipulated. 
These results increase the concerns that
computational complexity is indeed a barrier
to manipulation in practice. 



%
\bibliographystyle{ecai2010}
\bibliography{/Users/twalsh/Documents/biblio/a-z,/Users/twalsh/Documents/biblio/a-z2,/Users/twalsh/Documents/biblio/pub,/Users/twalsh/Documents/biblio/pub2}

\end{document}